\documentclass[%
 amsmath,amssymb,
superscriptaddress
]{revtex4-1}

\usepackage[utf8]{inputenc} % allow utf-8 input
\usepackage[T1]{fontenc}    % use 8-bit T1 fonts
\usepackage{hyperref}       % hyperlinks
\usepackage{url}            % simple URL typesetting
\usepackage{booktabs}       % professional-quality tables
\usepackage{amsfonts}       % blackboard math symbols
\usepackage{nicefrac}       % compact symbols for 1/2, etc.
\usepackage{microtype}      % microtypography

\usepackage{physics}

\usepackage{graphicx}

\usepackage{mathtools}

\usepackage{comment}

\usepackage{enumitem}

\usepackage{amssymb}

\usepackage{algorithm}
\usepackage{algorithmicx}
\PassOptionsToPackage{noend}{algpseudocode}% comment out if want end's to show
\usepackage{algpseudocode}

\makeatletter
\let\OldStatex\Statex
\renewcommand{\Statex}[1][3]{%
  \setlength\@tempdima{\algorithmicindent}%
  \OldStatex\hskip\dimexpr#1\@tempdima\relax}
\makeatother

\algnewcommand{\LineComment}[1]{\State \(\triangleright\) #1}

\newcommand{\ubf}{\boldsymbol{u}}

\newcommand{\thetabf}{\boldsymbol{\theta}}
\newcommand{\pt}{\tilde{p}}

\newcommand{\species}{{\cal R}}

\newcommand{\avep}[1]{\left \langle {#1} \right \rangle^{(d)}}
\newcommand{\avept}[1]{\left \langle {#1} \right \rangle^{(m)}}
\newcommand{\ave}[1]{\left \langle {#1} \right \rangle}

\newcommand{\lyr}[1]{^{({#1})}}

\newcommand{\adja}{\phi}
\newcommand{\adjW}{\Lambda}

\newcommand{\dkl}{{\cal D}_{\text{KL}}}

\newcommand{\momdiff}[1]{\Delta \mathbb{E} \left [ {#1} \right ]}

\begin{document}

\title{Deep Learning Moment Closure Approximations using Dynamic Boltzmann Distributions}
% \title{Dynamic Boltzmann Distributions for Deep Learning Moment Closure Approximations}

% Eric
% Dynamic Boltzmann distributions for spatial chemical reaction networks using deep Boltzmann machines / deep learning

%%%%%%%%%%%%%%%%%%%%%%%%%%%%%%%%%%%%%%%%
%%%%%%%%%%%%%%%%%%%%%%%%%%%%%%%%%%%%%%%%
% Authors
%%%%%%%%%%%%%%%%%%%%%%%%%%%%%%%%%%%%%%%%
%%%%%%%%%%%%%%%%%%%%%%%%%%%%%%%%%%%%%%%%

% The \author macro works with any number of authors. There are two commands
% used to separate the names and addresses of multiple authors: \And and \AND.
%
% Using \And between authors leaves it to LaTeX to determine where to break the
% lines. Using \AND forces a line break at that point. So, if LaTeX puts 3 of 4
% authors names on the first line, and the last on the second line, try using
% \AND instead of \And before the third author name.

\author{Oliver~K.~Ernst}
\affiliation{
  Department of Physics \\
  University of California, San Diego \\
  La Jolla, CA 92093 USA
}
\author{Tom~Bartol}
\affiliation{
  Salk Institute for Biological Studies \\
  La Jolla, CA 92037 USA
}
\author{Terrence~Sejnowski}
\affiliation{
  Salk Institute for Biological Studies \\
  La Jolla, CA 92037 USA
}
\author{Eric~Mjolsness}
\affiliation{
  Departments of Computer Science and Mathematics \\
  University of California, Irvine \\
  Irvine, CA 92697 USA
}

%%%%%%%%%%%%%%%%%%%%%%%%%%%%%%%%%%%%%%%%
%%%%%%%%%%%%%%%%%%%%%%%%%%%%%%%%%%%%%%%%
% Abstract
%%%%%%%%%%%%%%%%%%%%%%%%%%%%%%%%%%%%%%%%
%%%%%%%%%%%%%%%%%%%%%%%%%%%%%%%%%%%%%%%%

\begin{abstract}
% Infinite hierarchy
The moments of spatial probabilistic systems are often given by an infinite hierarchy of coupled differential equations. 
% Moment closure
Moment closure methods are used to approximate a subset of low order moments by terminating the hierarchy at some order and replacing higher order terms with functions of lower order ones.
% Problems
For a given system, it is not known beforehand which closure approximation is optimal, i.e. which higher order terms are relevant in the current regime. Further, the generalization of such approximations is typically poor, as higher order corrections may become relevant over long timescales.
% DBMs
We have developed a method to learn moment closure approximations directly from data using dynamic Boltzmann distributions (DBDs). 
% Finite elements
% The dynamics of the distribution are parameterized using basis functions from finite element methods such that the approach can be applied broadly to learn deep generative models in any application where systems of infinite differential equations arise.
The dynamics of the distribution are parameterized using basis functions from finite element methods, such that the approach can be applied without knowing the true dynamics of the system under consideration.
% Something about DBMs
We use the hierarchical architecture of deep Boltzmann machines (DBMs) with multinomial latent variables to learn closure approximations for progressively higher order spatial correlations.
% We show how a hierarchical architecture in the DBM and multinomial latent variables can be used to learn closure approximations for progressively higher order spatial correlations.
% emerging global statistics. 
% Centering
The learning algorithm uses a centering transformation, allowing the dynamic DBM to be trained without the need for pre-training.
% Lotka volterra
We demonstrate the method for a Lotka-Volterra system on a lattice, a typical example in spatial chemical reaction networks.
% Generalize
The approach can be applied broadly to learn deep generative models in applications where infinite systems of differential equations arise.
\end{abstract}

%%%%%%%%%%%%%%%%%%%%%%%%%%%%%%%%%%%%%%%%
%%%%%%%%%%%%%%%%%%%%%%%%%%%%%%%%%%%%%%%%
% Start document
%%%%%%%%%%%%%%%%%%%%%%%%%%%%%%%%%%%%%%%%
%%%%%%%%%%%%%%%%%%%%%%%%%%%%%%%%%%%%%%%%

\maketitle

%%%%%%%%%%%%%%%%%%%%%%%%%%%%%%%%%%%%%%%%
%%%%%%%%%%%%%%%%%%%%%%%%%%%%%%%%%%%%%%%%
% Introduction
%%%%%%%%%%%%%%%%%%%%%%%%%%%%%%%%%%%%%%%%
%%%%%%%%%%%%%%%%%%%%%%%%%%%%%%%%%%%%%%%%

\section{Introduction}

Infinite hierarchies of moment equations appear in many non-linear systems, including biochemical reaction networks~\cite{johnson_2015}, quantum optics~\cite{schack_1990},
and kinetic theory of gases~\cite{levermore_1996}, and more generally in the analysis of stochastic~\cite{kuehn_2015} and partial~\cite{frewer_2015} differential equations. 

Spatial reaction networks are a typical example of such systems. Differential equations for moments derived from the chemical master equation (CME)~\cite{gardiner_2009} usually depend on longer range spatial correlations. Moment closure approximations are widely used in chemical and biochemical modeling applications to obtain a solvable system by expressing higher order terms through lower order ones (see~\citet{johnson_2015} for a review on moment closure methods for reaction networks). 

Dynamic Boltzmann distributions (DBDs) have been introduced as generative models that learn a reduced dynamical system from data~\cite{ernst_2018_arxiv,ernst_2018,johnson_2015}. Here a differential equation model is introduced and learned, which can be parameterized either from the relevant physics, or otherwise using an appropriate set of basis functions. The learning problem can be formulated in continuous space in the form of a partial differential equation (PDE) constrained optimization problem. This offers a natural approach to introduce PDEs into machine learning. In the Supplemental Material, we review the general variational problem for DBDs, and make explicit the connection to the work in this paper.

The learning algorithm shares many advantages with the well-known Boltzmann machine learning algorithm~\cite{ackley_1985}. For example, in generative modeling approaches such as variational autoencoders (VAEs), the distribution over latent variables is chosen (typically Gaussian), while in DBDs it is learned from data. Further, the energy function in DBDs can be used to interpret the reduced models learned~\cite{ernst_2018}. Other approaches for learning temporal data such as recurrent neural networks (RNNs) do not offer a reduced model probability distribution, and do not give insight into the learned moment closure approximation as done in Section~\ref{sec:5}.

In this paper, we use the architecture of deep Boltzmann machines (DBMs) to train deep dynamic Boltzmann distributions for learning moment closure approximations. The learned model replaces long range spatial correlations with correlations involving latent variables, whose activations are learned from data. We show how the centering transformation~\cite{montavon_2012,melchior_2016} can be applied in the adjoint method~\cite{ernst_2018_arxiv,cao_2003} to derive a learning algorithm which does not require pre-training.

This paper is organized as follows: Section~2 introduces a simple example demonstrating moment hierarchies, Section~3 reviews DBMs and centered DBMs, Section~4 presents the learning algorithm for centered dynamic DBMs, and Section~5 presents numerical examples.

%%%%%%%%%%%%%%%%%%%%%%%%%%%%%%%%%%%%%%%%
%%%%%%%%%%%%%%%%%%%%%%%%%%%%%%%%%%%%%%%%
% Moment closure
%%%%%%%%%%%%%%%%%%%%%%%%%%%%%%%%%%%%%%%%
%%%%%%%%%%%%%%%%%%%%%%%%%%%%%%%%%%%%%%%%

\section{Moment closure in the Lotka-Volterra system} \label{sec:2}

%%%%%%%%%%%%%%%%%%%%%%%%%%%%%%%%%%%%%%%%
%%%%%%%%%%%%%%%%%%%%%%%%%%%%%%%%%%%%%%%%
% Figure 1
%%%%%%%%%%%%%%%%%%%%%%%%%%%%%%%%%%%%%%%%
%%%%%%%%%%%%%%%%%%%%%%%%%%%%%%%%%%%%%%%%

\begin{figure}[b]
	\centering
	\includegraphics[width=0.8\textwidth]{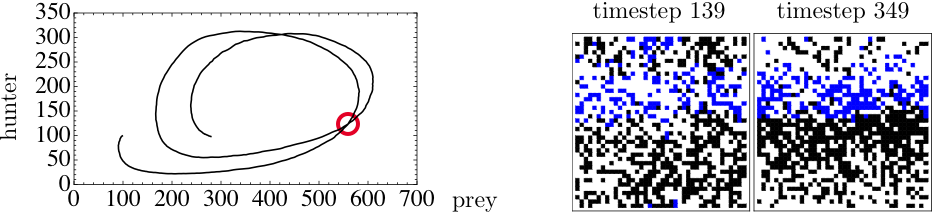}
	\caption{
	% Left
	\textit{Left:}
	Mean number of hunter and prey in the Lotka-Volterra system starting from $100$ particles of each population (see text).
	% Right
	\textit{Right:}
	Two lattices at an intersection point in the moment space of the left panel (red circle). Each have close to the average number of particles ($550$ prey in black and $140$ hunters in blue), but different spatial correlations lead to differing time derivatives~(\ref{eq:LVspatial}).
	}
	\label{fig:1}
\end{figure}

In this section, we review the appearance of a hierarchy of moments in a spatial probabilistic system. As an example, consider the Lotka-Volterra predator-prey system, described by the reactions:
%---------------
\begin{equation}
\varnothing \xrightarrow{k_1} P , \quad \quad \quad 
H \xrightarrow{k_2} \varnothing , \quad \quad \quad
H + P \xrightarrow{k_3} H + H ,
\end{equation}
%---------------
where $H$ and $P$ denote the predator (hunter) and prey populations and $k_1,k_2,k_3$ are reaction rates. 

In a well mixed system where spatial effects are not included, the mean number of $H$ and $P$, denoted by $\mu_H$ and $\mu_P$, obey the following ordinary differential equation (ODE) system:
%---------------
\begin{equation}
% \begin{split}
\frac{d \mu_P(t)}{dt} = k_1 \mu_P(t) - k_3 \mu_H(t) \mu_P(t)
\quad \& \quad
\frac{d \mu_H(t)}{dt} = k_3 \mu_H(t) \mu_P(t) - k_2 \mu_H(t) .
% \end{split}
\label{eq:LVhom}
\end{equation}
%---------------
These differential equations are solvable for given initial conditions $\mu_P(t=0)$ and $\mu_H(t=0)$.

In a spatial setting, consider a lattice in the single occupancy limit, where $s_{i,\alpha} \in \{ 0,1 \}$ describes the $i$-th lattice site occupied by species $\alpha \in \{H,P\}$. The mean number of $H$ and $P$ obey:
%---------------
\begin{equation}
\begin{split}
\frac{d}{dt} \ave{\sum_i s_{i,P} } &= k_1 \ave{\sum_i s_{i,P} } - k_3 \ave{\sum_{\langle ij \rangle} s_{i,H} s_{j,P} } , \\
\frac{d}{dt} \ave{\sum_i s_{i,H} } &= k_3 \ave{\sum_{\langle ij \rangle} s_{i,H} s_{j,P} } - k_2 \ave{\sum_i s_{i,H} } ,
\end{split}
\label{eq:LVspatial}
\end{equation}
%---------------
where $\sum_i$ sums over lattice sites, and $\sum_{\langle i j \rangle}$ over neighboring sites. These equations do not close - a nearest neighbor moment appears on the right hand side. Further, the differential equation describing this term depends on yet higher order ones, e.g. next-nearest neighbors, or three particle correlations (if particles are allowed to diffuse, the diffusion constant would appear here). Moment closure methods seek to obtain tractable approximations to this infinite hierarchy of differential equations.

\subsection{Stochastic simulations of reaction-diffusion systems} \label{sec:2.1}

Alternative to the deterministic approach of solving differential equations such as~(\ref{eq:LVhom}), stochastic approaches can be used. For example, the Gillespie stochastic simulation algorithm (SSA) can be derived from the CME~\cite{gillespie_1977,mjolsness_2013}. Spatial adaptations of the Gillespie SSA can similarly be used to generate stochastic simulations of~(\ref{eq:LVspatial}), including in continuous space~\cite{mcell_2}. Here we adopt a simple lattice based algorithm~\cite{ernst_2018_arxiv} in which particles hop on a grid, undergo unimolecular reactions following the Gillespie SSA, and bimolecular reactions with some chosen probability upon encounters.
%
\begin{comment}
A simple lattice based algorithm~\cite{takayasu_1992,ernst_2018_arxiv} we adopt in this work is as follows: iterate over all timesteps, at each step:
\begin{enumerate}
\item Perform unimolecular reactions following the standard Gillespie SSA.
\item Iterate over all particles in random order; for each:
\begin{enumerate}
\item Hop to a neighboring site, chosen at random with equal probability.
\item If the site is unoccupied, the move is accepted. If the site is occupied, a bimolecular reaction occurs with
some probability; else, the move is rejected and the particle is returned to the original site.
\end{enumerate}
\end{enumerate}
\end{comment}

To simulate the Lotka-Volterra system, we use a 2D lattice of $40 \times 40$ sites with von Neumann connectivity and periodic boundary conditions. For reaction rates, we use $k_1=0.025, k_3=0.06$, and bimolecular reaction probability $0.4$ corresponding to $k_2$. The initial state has $100$ particles of hunter and prey randomly distributed, and simulations are run for $500$ timesteps with $\Delta t = 1$. Figure~\ref{fig:1} shows snapshots of these simulations, which feature spatial patterns including waves.

We generate $100$ simulations as training data used in the rest of this paper. The mean number of hunter and prey is shown in Figure~\ref{fig:1}. Self intersections in this low order moment space reflect the dependence on spatial correlations which differ over time. 
% We use these simulations as training data for the rest of this paper.
The challenge for a deep learning problem is to identify relevant higher order correlations to separate states identical in low order moments, and to learn a closure model for expressing these correlations in terms of a finite number of parameters.

%%%%%%%%%%%%%%%%%%%%%%%%%%%%%%%%%%%%%%%%
%%%%%%%%%%%%%%%%%%%%%%%%%%%%%%%%%%%%%%%%
% DBMs for modeling reaction-diffusion systems
%%%%%%%%%%%%%%%%%%%%%%%%%%%%%%%%%%%%%%%%
%%%%%%%%%%%%%%%%%%%%%%%%%%%%%%%%%%%%%%%%

\section{DBMs for modeling reaction-diffusion systems}
    
Energy based models can be used to describe the state of reaction-diffusion system at an \textit{instant} in time. Here, we briefly review this notation~\cite{ernst_2018_arxiv} and key results for DBMs.

\subsection{Energy based models}

Let the lattice on which particles diffuse be the designated as the visible layer of the DBM. Let there be a total of $L$ layers, where $l=0$ denotes the visible layer and $l=1,\dots,L-1$ the hidden layers, each with $N\lyr{l}$ units. Let the units in the $l$-th layer be one of $M\lyr{l}$ species $\species\lyr{l} = \{ A, B, C, \dots \}$. % For example, for the hunter-prey system, the species in the visible layer are $\species\lyr{0} = \{ H, P \}$. Further, we choose to allow units in hidden layers to occupy one of $\species\lyr{l} = \{ H\lyr{l}, P\lyr{l} \}$ species.
The state of each layer is described by $N\lyr{l} \times M\lyr{l}$ a matrix, where entries $s_{i,\alpha} \in \{ 0, 1 \}$ denote the absence or presence of species $\alpha \in \species\lyr{l}$ at site $i = 1, \dots, N\lyr{l}$. We consider lattices in the single-occupancy limit, % such that each lattice site can only be occupied by at most one species, 
corresponding to the implicit constraint $\sum_{\alpha \in \species\lyr{l}} s_{i,\alpha}\lyr{l} \in \{ 0, 1 \}$. 

A general energy model for fully connected layers~\cite{ernst_2018_arxiv} has biases $a_{i,\alpha}\lyr{l}$ for unit $i$ in layer $l$ occupied by species $\alpha$, and weights $W_{i,\alpha,j,\beta}\lyr{l,l+1}$ connecting unit $i$ of species $\alpha$ in layer $l$ with unit $j$ of species $\beta$ in layer $l+1$. In this paper, we focus on learning hierarchical statistics with a smaller parameter space by making the following simplifications: 
%
% INLINE FORM
(1) we consider locally connected layers, where each unit in layer $l$ is connected to $q\lyr{l,l+1}$ units in layer $l+1$, and (2) biases and weights are shared across units in a layer: $a_{i,\alpha}\lyr{l} \rightarrow a_{\alpha}\lyr{l}$ and $W_{i,\alpha,j,\beta}\lyr{l,l+1} \rightarrow W_{\alpha\beta}\lyr{l,l+1}$. % (but remain species-dependent).
%
% LIST FORM
\begin{comment}
\begin{enumerate}[noitemsep,topsep=0pt]
\item We consider locally connected layers instead of fully connected, where each unit in layer $l$ is connected to $q\lyr{l,l+1}$ units in layer $l+1$.
\item Biases and weights are shared across units in a layer: $a_{i,\alpha}\lyr{l} \rightarrow a_{\alpha}\lyr{l}$ and $W_{i,\alpha,j,\beta}\lyr{l,l+1} \rightarrow W_{\alpha\beta}\lyr{l,l+1}$.
\end{enumerate}
Note that biases and weights remain species dependent.
\end{comment}
%
The energy function is: % of the DBM is:
%---------------
\begin{equation}
E(\{ \boldsymbol{S}\lyr{l} \}) = 
- \sum_{l=0}^{L-1} \sum_{\alpha \in \species\lyr{l}} a_\alpha\lyr{l} \sum_{i=1}^{N\lyr{l}} s_{\alpha,i}\lyr{l} 
- \sum_{l=0}^{L-2} \sum_{\alpha \in \species\lyr{l}} \sum_{\beta \in \species\lyr{l+1}} W_{\alpha \beta}\lyr{l,l+1} \sum_{\langle ij \rangle} s_{\alpha,i}\lyr{l} s_{\beta,j}\lyr{l+1} ,
\label{eq:energyNC}
\end{equation}
%---------------
where $\langle ij \rangle$ sums over the local connectivity between two layers. Since each site $i$ in layer $l$ is connected to $q\lyr{l,l+1}$ sites in layer $l+1$, then this sum comprises $N\lyr{l} \times q\lyr{l,l+1}$ terms in total.

\subsection{Learning rule for DBMs}

Maximizing the log likelihood of 
% the 
observed data 
% given the interaction parameters 
gives the well known learning rule for DBMs~\cite{salakhutdinov_2009}:
%---------------
\begin{equation}
\begin{split}
\Delta W_{\alpha\beta}\lyr{l,l+1} = \sum_{\langle ij \rangle} \momdiff{ s_{i,\alpha}\lyr{l} s_{j,\beta}\lyr{l+1} } \quad \quad &\& \quad \quad
\Delta a_\alpha\lyr{l} = \sum_{i=1}^{N\lyr{l}} \momdiff{ s_{i,\alpha}\lyr{l} } , % \\
% \text{where} \quad \momdiff{X} =& \avept{X} - \avep{X} ,
\end{split}
\end{equation}
%---------------
where $\momdiff{X} = \avept{X} - \avep{X}$, and $\avept{X}$ denotes an average taken over the model distribution, and $\avep{X}$ denotes an average taken over the data distribution, and the sign convention in the update steps is: $a_\alpha\lyr{l} \rightarrow a_\alpha\lyr{l} - \lambda \Delta a_\alpha\lyr{l}$ and similarly for $W_{\alpha\beta}\lyr{l,l+1}$, with learning rate $\lambda$.

Estimating the moments can be done using the well-known wake-sleep algorithm~\cite{ackley_1985}. The moments under the model distribution (sleep phase) are given by Gibbs sampling:
% Estimating the moments under the model distribution (sleep phase~\cite{ackley_1985}) can be done by Gibbs sampling:
% NEW ONE EQUATION
%---------------
\begin{equation}
\begin{split}
% Line 1
\pt(s_{i,\alpha}\lyr{l} = 1 | \{ \boldsymbol{S}\lyr{m \neq l} \} ) 
&= 
\exp [
\phi_{i,\alpha}\lyr{l}
]
\Big /
\Big (
1 + \sum_{\zeta \in \species\lyr{l}} \exp[ \phi_{i,\zeta}\lyr{l} ]
\Big ) , \\
% Line 2
\phi_{i,\alpha}\lyr{l} &= a_\alpha\lyr{l}
+ \sum_{\Delta l = \pm 1} \sum_{\beta \in \species\lyr{l+\Delta l}} W_{\alpha \beta}\lyr{l+\Delta l} \sum_{j \sim i} s_{\beta,j}\lyr{l+\Delta l} ,
\end{split}
\label{eq:gibbs}
\end{equation}
%---------------
% OLD TWO EQUATIONS
\begin{comment}
%---------------
\begin{equation}
\pt(s_{i,\alpha}\lyr{l} = 1 | \{ \boldsymbol{S}\lyr{m \neq l} \} ) 
= 
\exp [
\phi_{i,\alpha}\lyr{l}
]
\Big /
\Big (
1 + \sum_{\zeta \in \species\lyr{l}} \exp[ \phi_{i,\zeta}\lyr{l} ]
\Big ) ,
\label{eq:gibbs}
\end{equation}
%---------------
where the activation of species $\alpha$ at unit $i$ in layer $l$ is:
%---------------
\begin{equation}
\phi_{i,\alpha}\lyr{l} = a_\alpha\lyr{l}
+ \sum_{\Delta l = \pm 1} \sum_{\beta \in \species\lyr{l+\Delta l}} W_{\alpha \beta}\lyr{l+\Delta l} \sum_{j \sim i} s_{\beta,j}\lyr{l+\Delta l} ,
\end{equation}
%---------------
\end{comment}
% END COMMENT
%
where $\sum_{j \sim i}$ sums over units $j$ connected to unit $i$. Each step of sampling is performed in two phases: one pass for layers with even indexes, and one pass for layers with odd indexes. Estimating the moments under the data distribution (keeping the visible layer clamped at a data vector, i.e. the wake phase) can been done for DBMs by mean field methods~\cite{salakhutdinov_2009}, or else by Gibbs sampling~\cite{montavon_2012}.

\subsection{Centering transformation and the centered gradient}

A centered DBM~\cite{melchior_2016,montavon_2012} with parameters $\tilde{a}_\alpha\lyr{l}, W_{\alpha\beta}\lyr{l,l+1}$ has the energy function:
%---------------
\begin{equation}
\begin{split}
E(\{ \boldsymbol{S}\lyr{l} \}) = & 
- \sum_{l=0}^{L-1} \sum_{\alpha \in \species\lyr{l}} \tilde{a}_\alpha\lyr{l} \sum_{i=1}^{N\lyr{l}} \left ( s_{\alpha,i}\lyr{l} - \mu_\alpha\lyr{l} \right ) 
\\ &
- \sum_{l=0}^{L-2} \sum_{\alpha \in \species\lyr{l}} \sum_{\beta \in \species\lyr{l+1}} \tilde{W}_{\alpha \beta}\lyr{l,l+1} \sum_{\langle ij \rangle} \left ( s_{\alpha,i}\lyr{l} - \mu_\alpha\lyr{l} \right ) \left ( s_{\beta,j}\lyr{l+1} - \mu_\beta\lyr{l+1} \right ) ,
\end{split}
\end{equation}
%---------------
where $\mu_\alpha\lyr{l}$ are the species-dependent centers in layer $l$. Every regular DBM can be transformed to a centered DBM by transforming parameters as:
%---------------
\begin{equation}
\begin{split}
\tilde{W}_{\alpha\beta}\lyr{l,l+1} = W_{\alpha\beta}\lyr{l,l+1}  
\quad \quad  \& \quad \quad
\tilde{a}_{\alpha}\lyr{l} = a_\alpha\lyr{l} + \sum_{\Delta l = \pm 1} q\lyr{l,l+\Delta l} \sum_{\beta \in \species\lyr{l + \Delta l}} W_{\alpha\beta}\lyr{l,l+\Delta l} \mu_{\beta}\lyr{l + \Delta l} .
\end{split}
\label{eq:paramTransform}
\end{equation}
%---------------
This can be used to derive the \textit{centered gradient}~\cite{melchior_2016}: After sampling the moments of a regular DBM~(\ref{eq:gibbs}), transform to a centered DBM, calculate the gradient with respect to the centered parameters, and transform back to obtain the gradient for the regular DBM parameters. The result is
%---------------
\begin{equation}
\begin{split}
\Delta W_{\alpha\beta}\lyr{l,l+1} =& \sum_{\langle ij \rangle} \momdiff{ ( s_{i,\alpha}\lyr{l} - \mu_\alpha\lyr{l} ) ( s_{j,\beta}\lyr{l+1} - \mu_\beta\lyr{l+1} ) } , \\
\Delta a_\alpha\lyr{l} =& \sum_{i=1}^{N\lyr{l}} \momdiff{ s_{i,\alpha}\lyr{l} }
- \sum_{\Delta l = \pm 1} q\lyr{l,l+\Delta l} \sum_{\beta \in \species\lyr{l+\Delta l}} \Delta W_{\alpha\beta}\lyr{l,l+\Delta l} \mu_\beta\lyr{l+\Delta l} , % \\
% & \hspace{10mm} - q\lyr{l,l+1} \sum_{\beta \in \species\lyr{l+1}} \Delta W_{\alpha\beta}\lyr{l,l+1} \mu_\beta\lyr{l+1} ,
\end{split}
\end{equation}
%---------------
as derived in the Supplemental Material. To reduce noise, the centers are updated as the average unit's state with an exponential sliding average with sliding parameter $r \in [0,1]$:
%---------------
\begin{equation}
% \mu_\alpha^{(l,\text{new})} &= \frac{1}{N\lyr{l}} \sum_{i=1}^{N\lyr{l}} \avep{s_{i,\alpha}\lyr{l}} \\
\mu_\alpha\lyr{l} \leftarrow (1-r) \mu_\alpha\lyr{l} + r \times \frac{1}{N\lyr{l}} \sum_{i=1}^{N\lyr{l}} \avep{s_{i,\alpha}\lyr{l}} .
\label{eq:slide}
\end{equation}
%---------------

%%%%%%%%%%%%%%%%%%%%%%%%%%%%%%%%%%%%%%%%
%%%%%%%%%%%%%%%%%%%%%%%%%%%%%%%%%%%%%%%%
% Dynamic Centered DBMs
%%%%%%%%%%%%%%%%%%%%%%%%%%%%%%%%%%%%%%%%
%%%%%%%%%%%%%%%%%%%%%%%%%%%%%%%%%%%%%%%%

%%%%%%%%%%%%%%%%%%
%%%%%%%%%%%%%%%%%%

\section{Dynamic centered DBMs}

%%%%%%%%%%%%%%%%%%
%%%%%%%%%%%%%%%%%%

We follow earlier work on dynamic Boltzmann distributions~\cite{johnson_2015,ernst_2018,ernst_2018_arxiv} by escalating to time-varying interactions $a_\alpha\lyr{l}(t)$ and $W_{\alpha\beta}\lyr{l,l+1}(t)$ (see Supplemental Material for review). Instead of minimizing the Kullback-Leibler (KL) divergence of the model distribution $\pt$ to the data distribution $p$ at an instant in time, we now seek to minimize the KL divergence integrated over all times. Define as the objective function:
%---------------
\begin{equation}
S = \int_0^T dt \; \dkl(p || \tilde{p}) ,
\label{eq:objective}
\end{equation}
%---------------
% where $p$ is the data distribution, and $\tilde{p}$ is the dynamic Boltzmann distribution. 
%
The integral over time can lead to undesired extrema, particularly for periodic systems. In practice these can be eliminated using a small sliding time window:
$S = \int_{\tau}^{\tau + \Delta \tau} dt \; \dkl(p || \tilde{p})$, where $\Delta \tau$ is some small fixed value and $\tau$ is shifted forward every few optimization steps~\cite{ernst_2018_arxiv}.

\subsection{Reduced dynamic model}

To describe the time-evolving interactions, introduce the autonomous differential equation system:
%---------------
\begin{equation}
\begin{split}
\frac{d}{dt} a_\alpha\lyr{l}(t) &= F_{a_\alpha}\lyr{l} (\thetabf(t) ; \ubf_{a_\alpha}\lyr{l}) , \\
\frac{d}{dt} W_{\alpha\beta}\lyr{l,l+1}(t) &= F_{W_{\alpha\beta}}\lyr{l,l+1} (\thetabf(t) ; \ubf_{W_{\alpha\beta}}\lyr{l,l+1}) ,
\end{split}
\label{eq:constraints}
\end{equation}
%---------------
for given initial conditions $a_\alpha\lyr{l}(t=0)$, $W_{\alpha\beta}\lyr{l,l+1}(t=0)$. Here $\thetabf(t)$ is a chosen domain of interaction parameters (weights and biases), and $F$ are functions with free parameter vectors $\ubf$ to be learned. 
% Note that we will occasionally drop the arguments $F_{a_\alpha}\lyr{l} = F_{a_\alpha}\lyr{l} (\thetabf(t) ; \ubf_{a_\alpha}\lyr{l})$ to shorten notation.

The functions $F$ on the right side can be chosen based on the physics of the system under consideration to learn a reduced physical model. For reaction-diffusion systems, the forms of $F$ can be derived from the CME when using a fully visible Boltzmann distributions~\cite{ernst_2018}. A more blackbox aligned approach is to introduce basis functions $f_m(\thetabf(t))$ to parameterize~(\ref{eq:constraints}). Following~\citet{ernst_2018_arxiv} we use the $\mathbb{Q}_3$ family of finite elements~\cite{fem_table}, which has the advantage that in dimensions higher than one, basis functions are simply tensor products of 1D cubic polynomials $\mathbb{P}_3 \otimes \mathbb{P}_3 \otimes \dots$. The learning algorithm in Section~\ref{sec:4.3} requires $C_1$ functions - these polynomials are therefore the Hermite polynomials that in 1D control four degrees of freedom: the value and the first derivative at each endpoint. For $\thetabf$ of length $d$, this gives $4^d$ degrees of freedom in total and corresponding coefficients $\ubf$ to be learned:
%---------------
\begin{equation}
F_{a_\alpha}\lyr{l}(\thetabf(t) ; \ubf_{a_\alpha}\lyr{l}) = \sum_{m=1}^{4^d} u_{a_\alpha, m}\lyr{l} \times f_m (\thetabf(t)) 
\label{eq:basisFuncs}
\end{equation}
%---------------
and similarly for $W_{\alpha\beta}\lyr{l,l+1}$, where $f_m$ is the appropriate basis function.

%%%%%%%%%%%%%%%%%%
%%%%%%%%%%%%%%%%%%

\subsection{Moment closure approximation of dynamic centered DBMs}

%%%%%%%%%%%%%%%%%%
%%%%%%%%%%%%%%%%%%

The key advantage of the dynamic Boltzmann distribution setting is the moment closure approximation that can be learned from data: any given moment $\avept{X}$ evolves as (see Supplemental Material):
%---------------
\begin{equation}
\begin{split}
\frac{d \avept{X}}{dt} 
=& \sum_{l=0}^{L-1} \sum_{\alpha \in \species\lyr{l}} 
F_{a_\alpha}\lyr{l}(\thetabf (t); \ubf_{a_\alpha}\lyr{l}) \times \sum_{i=1}^{N\lyr{l}} \text{Cov} \left ( X, s_{i,\alpha}\lyr{l} \right ) \\
&+ \sum_{l=0}^{L-2} \sum_{\alpha \in \species\lyr{l}} \sum_{\beta \in \species\lyr{l+1}} 
F_{W_{\alpha\beta}}\lyr{l,l+1}(\thetabf (t); \ubf_{W_{\alpha\beta}}\lyr{l,l+1}) \times \sum_{\langle ij \rangle} \text{Cov} \left ( X, s_{i,\alpha}\lyr{l} s_{j,\beta}\lyr{l+1} \right ) ,
\end{split}
\label{eq:momentClosure}
\end{equation}
%---------------
where $\text{Cov}(X,Y) = \avept{XY} - \avept{X} \avept{Y}$.
The learned differential equation $F_\zeta$ of every interaction $\zeta$ (weights and biases) contributes to the closure model, weighted by a covariance term between $X$ and the observable for which $\zeta$ is the Lagrange multiplier. Equation~(\ref{eq:momentClosure}) should be directly compared against~(\ref{eq:LVspatial}). 
% Here, only first order terms in the visible layer appear on the right hand side. 
Higher order terms of visible units appearing in~(\ref{eq:LVspatial}) are exchanged for correlations with latent random variables, whose activations are \textit{learned}. % to capture relevant observables.

%%%%%%%%%%%%%%%%%%
%%%%%%%%%%%%%%%%%%

\subsection{Adjoint method learning problem with centering transformation} \label{sec:4.3}

%%%%%%%%%%%%%%%%%%
%%%%%%%%%%%%%%%%%%

We next formulate a learning problem for the parameters $\ubf$ defining the dynamical system~(\ref{eq:constraints}). This is a specific case of a variational problem for the functions appearing on the right hand side of a differential equation~\cite{ernst_2018_arxiv,gamkrelidze_book}, as shown in the Supplemental Material. To enforce the constraints~(\ref{eq:constraints}), introduce adjoint variables $\adja_\alpha\lyr{l},\adjW_{\alpha\beta}\lyr{l,l+1}$ to the \textit{centered} parameters $\tilde{a}_\alpha\lyr{l}, \tilde{W}_{\alpha\beta}\lyr{l,l+1}$.
The adjoint equations can be derived from the Hamiltonian (Supplemental Material) giving
% The result is
%---------------
\begin{equation}
\begin{split}
% a
\frac{d}{dt} \adja_\alpha\lyr{l} =& 
% a - kl div
\sum_{i=1}^{N\lyr{l}} \momdiff{ s_{i,\alpha}\lyr{l} }
% a - psi term
- \psi_{a_\alpha\lyr{l}} , \\
% W
\frac{d}{dt} \adjW_{\alpha\beta}\lyr{l,l+1} =& 
% W - kl div
\sum_{\langle ij \rangle} \momdiff{ (s_{i,\alpha}\lyr{l} - \mu_\alpha\lyr{l}) (s_{j,\beta}\lyr{l+1} - \mu_\beta\lyr{l+1}) }
% W - psi terms
- \psi_{W_{\alpha\beta}\lyr{l,l+1}} \\
& + q\lyr{l,l+1} \left ( \mu_\alpha\lyr{l} \psi_{a_\beta\lyr{l+1}}
+ \mu_\beta\lyr{l+1} \psi_{a_\alpha\lyr{l}}
% W - dmu/dt term: m-1 -> l and m -> l+1 and gamma -> beta and delta -> alpha
- \adja_\beta\lyr{l+1} \frac{d \mu_\alpha\lyr{l}}{dt}
% W - dmu/dt term: m -> l and m+1 -> l+1 and gamma -> alpha and delta -> beta
- \adja_\alpha\lyr{l} \frac{d \mu_\beta\lyr{l+1}}{dt} 
\right ) ,  \\
% PSI
% \psi_\theta = & 
% a - deriv W
% \sum_{m=0}^{L-2} \sum_{\zeta \in \species\lyr{m}} \sum_{\eta \in \species\lyr{m+1}} \adjW_{\zeta\eta}\lyr{m,m+1} \frac{\partial F_{W_{\zeta\eta}}\lyr{m,m+1}}{\partial \theta}
% a - start adjoint
% + \sum_{m=0}^{L-1} \sum_{\zeta \in \species\lyr{m}} \adja_\zeta\lyr{m} \times \\ 
% & \Bigg (
% a - adjoint - deriv a
% \frac{\partial F_{a_\zeta}\lyr{m}}{\partial \theta}
% a - adjoint - deriv l-1
%+ \sum_{\Delta m = \pm 1} q\lyr{m,m+\Delta m} \sum_{\eta \in \species\lyr{m+\Delta m}} \frac{\partial F_{W_{\zeta\eta}}\lyr{m,m+\Delta m}}{\partial \theta} \mu_\eta\lyr{m+\Delta m}
% \Bigg ) ,
\end{split}
\label{eq:adjoint}
\end{equation}
%---------------
with boundary conditions $\adja_\alpha\lyr{l}(t=\tau+\Delta\tau) = \adjW_{\alpha\beta}\lyr{l,l+1}(t=\tau+\Delta\tau) = 0$, and where
%---------------
\begin{equation}
\begin{split}
\psi_\theta = & 
% a - deriv W
\sum_{m=0}^{L-2} \sum_{\zeta \in \species\lyr{m}} \sum_{\eta \in \species\lyr{m+1}} \adjW_{\zeta\eta}\lyr{m,m+1} \frac{\partial F_{W_{\zeta\eta}}\lyr{m,m+1}}{\partial \theta}
% a - start adjoint
+ \sum_{m=0}^{L-1} \sum_{\zeta \in \species\lyr{m}} \adja_\zeta\lyr{m} \times \\ 
& \Bigg (
% a - adjoint - deriv a
\frac{\partial F_{a_\zeta}\lyr{m}}{\partial \theta}
% a - adjoint - deriv l-1
+ \sum_{\Delta m = \pm 1} q\lyr{m,m+\Delta m} \sum_{\eta \in \species\lyr{m+\Delta m}} \frac{\partial F_{W_{\zeta\eta}}\lyr{m,m+\Delta m}}{\partial \theta} \mu_\eta\lyr{m+\Delta m}
% a - adjoint - deriv l+1
% + q\lyr{m,m+1} \sum_{\eta \in \species\lyr{m+1}} \frac{\partial F_{W_{\zeta\eta}}\lyr{m,m+1}}{\partial \theta} \mu_\eta\lyr{m+1} 
\Bigg ) .
\end{split}
\label{eq:psi}
\end{equation}
%---------------
While analytic expression for $d \mu_\alpha\lyr{l} / dt$ are not available, in practice they can be easily estimated as: 
% between values obtained by sampling at successive timesteps: 
$d \mu_\alpha\lyr{l}(t) / dt \approx ( \mu_\alpha\lyr{l}(t+\Delta t) - \mu_\alpha\lyr{l}(t) ) / \Delta t$.
%
% Similarly to the centered gradient, we note that the non-centered version of the adjoint system~(\ref{eq:adjoint}) is directly obtained by taking $\mu_\alpha\lyr{l} \rightarrow 0$.
%

The sensitivity (update) equations for the parameters $\ubf$ to be learned are:
%---------------
\begin{equation}
\begin{split}
\frac{dS}{d \ubf_{a_\alpha}\lyr{l}} =& - \int_\tau^{\tau+\Delta\tau} dt \; \adja_\alpha\lyr{l} \frac{\partial F_{a_\alpha}\lyr{l}}{\partial \ubf_{a_\alpha}\lyr{l}} , \\
\frac{dS}{d \ubf_{W_{\alpha\beta}}\lyr{l,l+1}} =& - \int_\tau^{\tau+\Delta\tau} dt \; \left ( \adjW_{\alpha\beta}\lyr{l,l+1} + q\lyr{l,l+1} \adja_\alpha\lyr{l} \mu_\beta\lyr{l+1} + q\lyr{l,l+1} \adja_\beta\lyr{l+1} \mu_\alpha\lyr{l} \right ) \frac{\partial F_{W_{\alpha\beta}}\lyr{l,l+1}}{\partial \ubf_{W_{\alpha\beta}}\lyr{l,l+1}} ,
\end{split}
\label{eq:sensitivity2}
\end{equation}
%---------------
where the update step is: $\ubf \rightarrow \ubf - \lambda \times ( dS / d \ubf )$ with learning rate $\lambda$ (see Supplemental Material).

Algorithm~1 is an example of how the learning problem can be solved in practice. Alternative approaches for solving PDE-constrained optimization problems can be applied, such as sequential quadratic programming (SQP). A benefit of the current algorithm is its simplicity - the inner loop at each timestep is equivalent to the wake/sleep phases of the Boltzmann machine learning algorithm~\cite{ackley_1985}. 
% Note that non-centered parameters are used for sampling, as in the centered gradient. 
A lower bound on the log-probability of test data can therefore be obtained using established methods such as Annealed Importance Sampling (AIS)~\cite{salakhutdinov_2009}.
It is naturally possible to use accelerated gradient descent methods such as Adam~\cite{adam}. 

%%%%%%%%%%%%%%%%%%%%%%%%%%%%%%%%%%%%%%%%
%%%%%%%%%%%%%%%%%%%%%%%%%%%%%%%%%%%%%%%%
% Algorithm
%%%%%%%%%%%%%%%%%%%%%%%%%%%%%%%%%%%%%%%%
%%%%%%%%%%%%%%%%%%%%%%%%%%%%%%%%%%%%%%%%

\begin{algorithm}[H]
\caption{Learning algorithm for dynamic centered DBMs} \label{alg:1}
\begin{algorithmic}[1]
\State \textbf{Input:} Initial conditions for all interactions parameters, time interval $[0,T]$, a formula for the learning rate $\lambda$, sliding factor $r$, batch size $\eta$, time window size $\Delta \tau$, a formula for sliding $\tau$.
\State \textbf{Initialize:} $\tau=0$, $\mu_\alpha\lyr{0}$ to the data means, otherwise $\mu_\alpha\lyr{l} = 1 / (M\lyr{l} + 1)$.
\While{not converged}
    \For{timepoint $t$ in $[\tau,\tau+\Delta\tau]$ with timestep $\Delta t$}
        \LineComment{Solve the constraints~(\ref{eq:constraints}) for the current timepoint $t$ from the previous timepoint $t - \Delta t$.}
        \LineComment{Estimate wake \& sleep phase moments~(\ref{eq:adjoint}) over the batch (e.g. Gibbs sampling).}
        \LineComment{Update the centers $\mu_\alpha\lyr{l}(t)$ according to~(\ref{eq:slide}).}
        \LineComment{Calculate \& store derivative terms $\partial F / \partial \theta$ in~(\ref{eq:adjoint}) and $\partial F / \partial \ubf$ in~(\ref{eq:sensitivity2}).}
	\EndFor
	\LineComment{Solve the adjoint system~(\ref{eq:adjoint}) backwards in time from $t=\tau+\Delta \tau$ to $t=\tau$.}
	\LineComment{Update the parameters $\ubf_{a_\alpha}\lyr{l}$ and $\ubf_{W_{\alpha\beta}}\lyr{l,l+1}$ according to~(\ref{eq:sensitivity2}) with learning rate $\lambda$.}
	\LineComment{Slide the time window at $\tau$ forward if necessary to eventually cover $[0,T]$.}
\EndWhile
\end{algorithmic}
\end{algorithm}

\section{Numerical examples} \label{sec:5}

We demonstrate Algorithm~1 for learning a moment closure approximation for the Lotka-Volterra system of Section~\ref{sec:2}. As training data, we use the stochastic simulations generated in Section~\ref{sec:2.1}. Note that we only consider a single initial condition in this work - however, it is possible to learn a larger parameter space from stochastic simulations with varying initial conditions~\cite{ernst_2018_arxiv}.

\begin{figure}[t]
	\centering
	\includegraphics[width=1.0\textwidth]{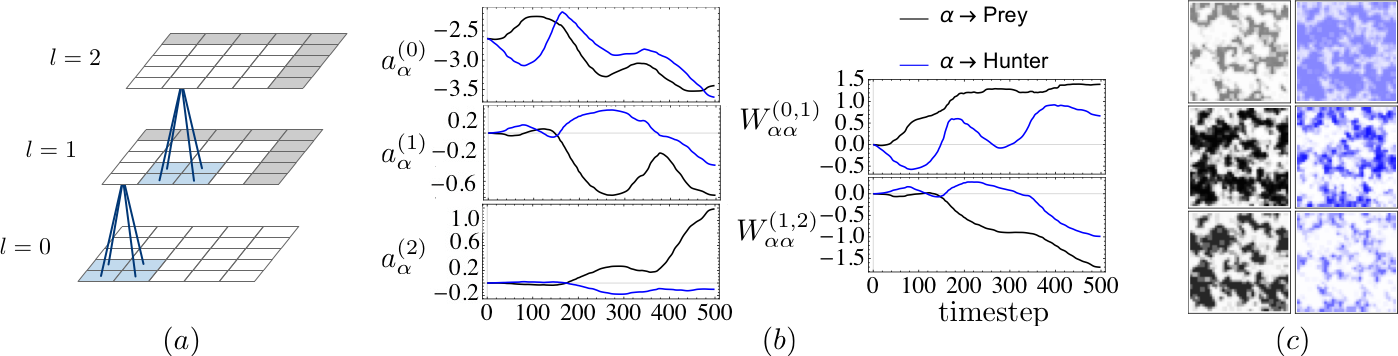}
	\caption{
	(a)
	Locally connected DBM in 3 layers with $40\times40$ units each ($5\times5$ illustrated). Every $2\times2$ patch of units in layer $l$ (blue) is connected to a single unit in layer $l+1$ with two species-dependent weights $W_{HH}\lyr{l,l+1}$ and $W_{PP}\lyr{l,l+1}$. Gray units implement periodic boundary conditions to the layer below. % (stochastic simulations also use periodic boundaries).
	(b)
	Interactions learned for the Lotka-Volterra system using Algorithm~1 (see text).
	(c)
	Spatial patterns in the layers of (a) at timepoint $370$ after training, obtained by $100$ steps of Gibbs sampling from a random configuration (raw probabilities shown for prey in black, hunter in blue).
	% for hunter and prey populations are shown, giving spatial patterns visually similar to the stochastic simulations. 
	}
	\label{fig:2}
\end{figure}

For the architecture of the DBM, we use a locally connected DBM with one visible and two hidden layers as shown in Figure~\ref{fig:2}(a). Each $2\times2$ patch in layer $l$ is connected to a single unit in layer $l+1$. The number of units in each layer is held constant at $40\times40$, with boundary units implementing periodic boundary conditions to the layer below, reflecting those used in the stochastic simulations.

Many systems feature a large number of species, e.g. states of ion channels as different sub-units are activated. Having $M\lyr{l}$ species in layer $l$ and $M\lyr{l+1}$ species in layer $l+1$ leads to $M\lyr{l} \times M\lyr{l+1}$ species dependent weights. To limit this parameter inflation, we consider two species $H,P$ in each layer, and weights $W_{HH}\lyr{0,1}, W_{PP}\lyr{0,1}$ and $W_{HH}\lyr{1,2}, W_{PP}\lyr{1,2}$. We found the omitted cross-species weights to be less important, as a similar effect is produced by crowding in the hidden layers during sampling.

\begin{figure}[b]
	\centering
	\includegraphics[width=1.0\textwidth]{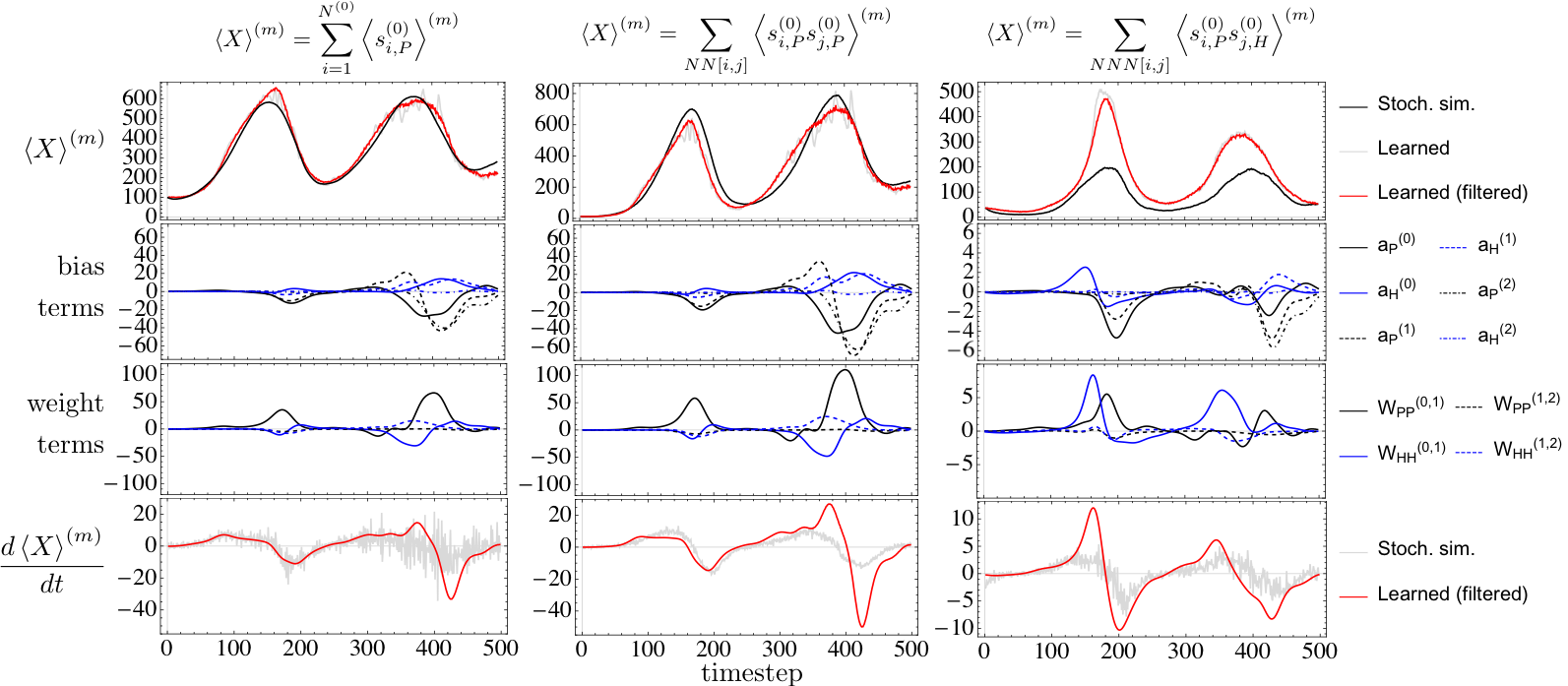}
	\caption{\textit{Columns:} Quantities of interest $\avept{X}$. \textit{Top row:} Comparison with stochastic simulations. \textit{Second, third:} Terms in the moment closure approximation~(\ref{eq:momentClosure}). \textit{Bottom:} Derivative from summing the second \& third rows as in~(\ref{eq:momentClosure}) compared to the derivative from stochastic simulations.}
	\label{fig:3}
\end{figure}

As initial condition, we use the maximum entropy distribution consistent with the $100$ initial hunter and prey particles. This corresponds to initial parameters $a_H\lyr{0} = a_P\lyr{0} = -2.63$, and all other weights and biases set to zero. The algorithm therefore must learn to activate latent variables to control spatial correlations. Starting from zero weights and biases allows the hidden layers to enforce either a strengthening or suppressing (positive or negative weights) relationship with neighboring layers.
We restrict the domain of the differential equations~(\ref{eq:constraints}) to be $\thetabf~=~(a_H\lyr{0},a_P\lyr{0},W_{HH}\lyr{0,1},W_{PP}\lyr{0,1},W_{HH}\lyr{1,2},W_{PP}\lyr{1,2})$. Since the distribution starts with only the visible biases as non-zero, it is important to include these in the domain. % , as the other parameters initially vary only slowly. 
When training on stochastic simulations from many initial conditions, using fewer parameters in the domain improves generalization~\cite{ernst_2018_arxiv}. The side length of the cubic cell~(\ref{eq:basisFuncs}) used in all dimensions was $0.1$.

Algorithm~1 is implemented as a C++ library freely available online~\cite{dblz} (included in the Supplemental Material). We use simple batch gradient descent with batch size $\eta = 5$, learning rate $\lambda = 2.5 \times 10^{-6}$ for all parameters for $1000$ optimization steps, and sliding factor $r=0.5$ (the center is averaged over all units allowing a larger factor than in non-dynamic centered DBMs). We use $10$ steps of Gibbs sampling for estimating both the wake and sleep phase moments, with persistent chains in the sleep phase~\cite{tieleman_2008}. The differential equations~(\ref{eq:constraints},\ref{eq:adjoint}) are solved using Euler's method with the step size from the stochastic simulations. The time window in~(\ref{eq:sensitivity2}) is of size $\Delta \tau = 10$ timesteps, and we slide $\tau \rightarrow \tau+1$ every two optimization steps. Figure~\ref{fig:2}(b) shows the learned parameter trajectories. An example of the hidden layer states is shown in Figure~\ref{fig:2}(c), showing the learned hierarchical representation of spatial patterns in the hidden layers.

The moment closure approximation~(\ref{eq:momentClosure}) is of particular interest. Here it can be analyzed which weighted covariance terms the system has learned that contribute to the time evolution of an observable. Consider for example the mean number of prey $\avept{X} = \sum_{i=1}^{N\lyr{0}} \big \langle s_{i,\alpha}\lyr{0} \big \rangle^{(m)}$. Figure~\ref{fig:3}, top-left, shows the time evolution of this observable under the trained model obtained by averaging over $100$ sampled lattices ($10$ steps of Gibbs sampling from a random configuration) using the learned parameters at each timestep. For testing, we compare the trajectory against stochastic simulations, which agree well (for training on varying initial conditions, a test data set of stochastic simulations may be used~\cite{ernst_2018_arxiv}). Next, we calculate each term in~(\ref{eq:momentClosure}), shown in rows two and three (to reduce noise, we first smooth the interactions with a low-pass filter with cutoff $0.1$ before sampling and calculating these terms). The sign of weight and bias terms is generally opposite - this reflects the selective activation of units based on the $2\times2$ patches in neighboring layers, rather than broad, spatially uncorrelated activations. To validate the accuracy of these terms, we also plot their sum in the bottom row, which is nearly identical to the true time derivative $d \avep{X} / dt$ calculated from the stochastic simulations.

The two oscillation cycles in Figure~\ref{fig:1} evolve differently due to the dependence on higher order moments. How is this difference reflected in the learned moment closure approximation? In the first oscillation in Figure~\ref{fig:3}, the time evolution is primarily driven by the weight term $W_{PP}\lyr{0,1}$ from the first hidden layer, indicating that mainly nearest neighbor structure relevant. In the second oscillation, the contribution from terms in the second hidden layer has grown significantly, indicating longer range correlations are relevant. Indeed, visually inspecting samples of the stochastic simulations (see Figure~\ref{fig:1}) shows that larger domains of hunter and prey are formed in the second oscillation.

Examining different quantities $\avept{X}$ in this fashion gives insight into the learned moment closure approximation. In the center column of Figure~\ref{fig:3}, we examine the mean number of nearest neighbors of prey. The terms in~(\ref{eq:momentClosure}) are approximately scaled version of the terms for the mean number of prey. This is expected, since the reaction system does not explicitly give a source or sink for nearest neighbor correlations between prey. Not all terms are accurately captured. For example, the right column shows the mean number of next-nearest neighbors (Manhattan distance two) of hunter and prey, which are overestimated in the learned model. This is explained by weight terms from hunter and prey both contributing positively to the time evolution, rather than competitively as previously.

%%%%%%%%%%%%%%%%%%%%%%%%%%%%%%%%%%%%%%%%
%%%%%%%%%%%%%%%%%%%%%%%%%%%%%%%%%%%%%%%%
% Discussion
%%%%%%%%%%%%%%%%%%%%%%%%%%%%%%%%%%%%%%%%
%%%%%%%%%%%%%%%%%%%%%%%%%%%%%%%%%%%%%%%%

\section{Discussion}

We have introduced a method for learning moment closure approximations from data using multiple hidden layers. A key result is the closure equation~(\ref{eq:momentClosure}), which replaces long range spatial correlations in the visible layer with correlations with latent variables, whose activation is learned. The learning problem is that for a dynamic Boltzmann distribution, combined with the architecture of a DBM. The centering transformation from centered DBMs is extended to the adjoint system for the dynamic case, such that pre-training is unnecessary. The hierarchical architecture in Figure~\ref{fig:2}(a) is tailored to reflect the moment equations derived CME~(\ref{eq:LVspatial}), naturally capturing correlations relevant to the moment closure problem. A further important result is the use of multinomial variables in the hidden layers, which allows interpretable learned representations as in Figure~\ref{fig:2}(c).

Further avenues for improvement exist. It may be possible to adapt the ``serendipitous" family of $\mathbb{Q}_3$ finite elements~\cite{fem_table} to reduce the number of basis functions and therefore computational cost~\cite{holst_2015}, although it is not $C_1$, requiring modifications to the learning problem. For modeling applications, physically informed parameterizations are particularly interesting, e.g. for reaction-diffusion systems~\cite{ernst_2018_arxiv}, and generally in mathematical modeling~\cite{mjolsness_2018}.

%%%%%%%%%%%%%%%%%%%%%%%%%%%%%%%%%%%%%%%%
%%%%%%%%%%%%%%%%%%%%%%%%%%%%%%%%%%%%%%%%
% Acknowledgments
%%%%%%%%%%%%%%%%%%%%%%%%%%%%%%%%%%%%%%%%
%%%%%%%%%%%%%%%%%%%%%%%%%%%%%%%%%%%%%%%%

\begin{acknowledgements}

This work was supported by 
% All of us
National Institute of Aging grant R56AG059602 (E.M., O.K.E., T.B., T.S.), 
% Eric
Human Frontiers Science Program grant HFSP - RGP0023/2018 (E.M.),
% Tom Terry & Me?
and NIH P41-GM103712, NIH R01MH115456, and AFOSR MURI FA9550-18-1-0051 (O.K.E., T.B., T.S.).

\end{acknowledgements}

%%%%%%%%%%%%%%%%%%%%%%%%%%%%%%%%%%%%%%%%
%%%%%%%%%%%%%%%%%%%%%%%%%%%%%%%%%%%%%%%%
% References
%%%%%%%%%%%%%%%%%%%%%%%%%%%%%%%%%%%%%%%%
%%%%%%%%%%%%%%%%%%%%%%%%%%%%%%%%%%%%%%%%

% \section*{References}

\bibliographystyle{plainnat_custom}
\bibliography{bibliography}

\end{document}